\documentclass{llncs}

\usepackage{amsmath}
\usepackage{amssymb}
\usepackage{bussproofs}
\usepackage[caption=false,position=top]{subfig}
\usepackage{listings}
\usepackage{enumerate}
\usepackage{graphicx}
\usepackage[binary-units=true]{siunitx}
\usepackage{todonotes}
\usepackage{hyperref}
\usepackage{pgfplots}
\pgfplotsset{compat=1.3}
\usetikzlibrary{patterns}
\usepgfplotslibrary{external}

\title{The Higher-Order Prover Leo-III\thanks{The work is supported by
    the German National Research Foundation (DFG) under grant BE 2501/11-1 (LEO-III).}\\(Extended Version)}
\titlerunning{The Higher-Order Prover Leo-III}
\author{Alexander Steen\inst{1} \and Christoph Benzm\"uller\inst{2, 1}}
\authorrunning{Steen \and Benzm\"uller}
\institute{Freie Universit\"at Berlin, Institute of Computer Science, Berlin, Germany
\and University of Luxembourg, Computer Science and Communications,
Luxembourg\\
\email{\{a.steen,c.benzmueller\}@fu-berlin.de}}

\def\sys{\mbox{\textsc{LeoPARD}}}

\definecolor{blue_1}{RGB}{0,90,169}
\definecolor{blue_2}{RGB}{0,131,204}
\definecolor{blue_1b}{RGB}{93,133,195}
\definecolor{blue_2b}{RGB}{0,156,218}
\definecolor{gray_1}{RGB}{142,152,178}

\begin{document}
\maketitle

\begin{abstract}
  The automated theorem prover Leo-III for classical higher-order
  logic with Henkin semantics and choice is presented. Leo-III is
  based on extensional higher-order paramodulation and accepts every
  common TPTP dialect (FOF, TFF, THF), including their recent
  extensions to rank-1 polymorphism (TF1, TH1). In addition, the
  prover natively supports almost every  normal higher-order
  modal logic. Leo-III cooperates with first-order reasoning tools
  using translations to many-sorted first-order logic
  and produces verifiable proof certificates. The prover is evaluated
  on heterogeneous benchmark sets. 
\end{abstract}

\newcommand{\semtrue}{\text{T}}
\newcommand{\semfalse}{\text{F}}
\newcommand{\syntrue}{\top}
\newcommand{\synfalse}{\bot}
\newcommand{\clause}[1]{{\mathcal{#1}}}
\newcommand{\pospol}{\mathfrak{t\!t}}
\newcommand{\negpol}{\mathfrak{f\!f}}
\newcommand{\hd}{\mathit{hd}}
\newcommand{\fv}{\mathit{fv}}

\section{Introduction}
Leo-III is an automated theorem prover (ATP) for classical higher-order logic (HOL)
with Henkin semantics and choice.\footnote{Leo-III is freely available (BSD license) at \url{http://github.com/leoprover/Leo-III}.} It is the successor of the
well-known LEO-II prover~\cite{J30}, whose development significantly
influenced the build-up of the TPTP THF infrastructure~\cite{J22}.
Leo-III exemplarily utilizes and instantiates the associated~\sys~
system platform~\cite{C45short} for higher-order (HO) deduction systems
implemented in Scala.

In the tradition 
of the cooperative nature of the LEO prover family, Leo-III
collaborates with external theorem
provers during proof search, in particular,
with first-order (FO) ATPs such as E~\cite{Schulz:2002:EBT:1218615.1218621}.
Unlike LEO-II,
which translated proof obligations into untyped FO languages,
Leo-III, by default, translates its HO clauses to
(monomorphic or polymorphic) many-sorted FO formulas. That
way clutter is reduced during translation, resulting in a more
effective cooperation.


Leo-III supports all common
TPTP~\cite{DBLP:journals/jar/Sutcliffe17,J22} dialects (CNF, FOF, TFF,
THF) as well as the polymorphic variants TF1 and
TH1~\cite{DBLP:conf/cade/BlanchetteP13,DBLP:conf/cade/KaliszykSR16}.
The prover returns results according to the standardized SZS ontology
and additionally produces a TSTP-compatible (refutation) proof
certificate, if a proof is found. Furthermore, Leo-III
natively supports reasoning for almost every normal HO modal
logic, including (but not limited to)
logics \textbf{K}, \textbf{D}, \textbf{T}, \textbf{S4} and \textbf{S5} with
constant, cumulative or domain quantifiers~\cite{blackburn2006handbook}.  These hybrid logic
competencies make Leo-III, up to the authors' knowledge, the most
widely applicable ATP 
available to date.

The most current release of Leo-III (version 1.2) comes with several 
novel features, including specialized calculus rules for function
synthesis, injective functions and equality-based
simplification. An evaluation of Leo-III 1.2 confirms that
it is on a par with other current state-of-the-art HO ATP systems.

This paper outlines the base calculus of Leo-III and 
highlights the novel features of version 1.2. As a pioneering
contribution this also includes Leo-III's native support for reasoning
in HO modal logics (which, of course, includes propositional
and FO modal logics). Finally, an evaluation of Leo-III is presented
for all monomorphic and polymorphic THF problems from the TPTP
problem library and for all mono-modal logic problems from the QMLTP
library~\cite{qmltp}.


\textit{Related ATP systems.} 
These include
TPS, Satallax, cocATP and
agsyHOL.  Also, some interactive proof assistants
such as Isabelle/HOL can be used for
automated reasoning in HOL.  More weakly related systems include the
various recent
attempts to lift FO ATPs to the HO domain, e.g. Zipperposition.

\textit{Higher-Order Logic.}
HOL as addressed here has been proposed by Church, and further studied
by Henkin, Andrews and others, cf.~\cite{sep-type-theory-church,B5}
and the references therein. 
  It provides lambda-notation, as an
 elegant and useful means to denote unnamed functions, predicates and
 sets (by their characteristic functions). 
In the remainder a notion of
HOL with Henkin semantics and choice is assumed.


\section{Higher-Order Paramodulation}    
 
  Leo-III extends a complete, paramodulation based calculus for HOL
            with practically motivated, heuristic inference
            rules, cf.~Fig.~\ref{fig:calculus}. They are grouped as follows:
  
  \begin{figure}[t] 
      \fbox{
        \begin{minipage}{\textwidth}
          \fbox{\textsc{\scriptsize Primary inferences}} \\
          \begin{minipage}{.49\textwidth}
          \begin{prooftree}
             \AxiomC{$\clause{C} \lor [l \simeq r]^\pospol$}
             \AxiomC{$\clause{D} \lor [s \simeq t]^\alpha$}
             \RightLabel{\scriptsize(Para)}
             \BinaryInfC{$\clause{C} \lor \clause{D} \lor
                  [s[r]_\pi \simeq t]^\alpha \lor [s |_\pi \simeq l]^\negpol$}
          \end{prooftree}
          \end{minipage}
          \begin{minipage}{.49\textwidth}
          \begin{prooftree}
             \AxiomC{$\clause{C} \lor [l \simeq r]^\alpha \lor [s \simeq t]^\alpha$}
             \RightLabel{\scriptsize(EqFac)}
             \UnaryInfC{$\clause{C} \lor [l \simeq r]^\alpha \lor [l \simeq s]^\negpol
                          \lor [r \simeq t]^\negpol$}
          \end{prooftree}
          \end{minipage}
          \begin{prooftree}
           \AxiomC{$\clause{C} \lor [X_{\overline{\tau_i}\to o} \; \overline{t^i_{\tau_i}}]^\alpha$}
           \AxiomC{$p \in \mathcal{GB}^{\{\neg,\lor\} \cup \{\Pi^\tau\!,\, ={}^\tau \mid \tau \in \mathcal{T}\}}_{\overline{\tau_i}\to o}$}
           \RightLabel{\scriptsize(PS)}
           \BinaryInfC{$\big(\clause{C} \lor
              [X_{\overline{\tau_i}\to o} \; \overline{t^i_{\tau_i}}]^\alpha\big)\{X / p\}$}
          \end{prooftree}
          \vspace{-.5em}
          \hrulefill \\
          \fbox{\textsc{\scriptsize Further rules}} \raisebox{1pt}{\tiny
          \begin{minipage}{.8\textwidth}
          (Choice): $E$ is a choice operator or a free variable; X is a fresh variable.\\
          (INJ): $\mathrm{sk}$ is a fresh constant symbol of appropriate type.
          \end{minipage}
          }
          \\
          \begin{minipage}{.49\textwidth}
          \begin{prooftree}
    \AxiomC{$\clause{C} \lor [s[E \; t]]^\alpha$}
    \RightLabel{\scriptsize (Choice)}
    \UnaryInfC{$[t \; X]^\negpol \lor [t \; (\epsilon \; t)]^\pospol$}
  \end{prooftree}
          \end{minipage}
          \begin{minipage}{.4\textwidth}
          \begin{prooftree}
    \AxiomC{$[f \; X \simeq f \; Y]^\pospol \lor [X \simeq Y]^\negpol$}
    \RightLabel{\scriptsize (INJ)}
    \UnaryInfC{$[\mathrm{sk} \; (f \; X) \simeq X]^\pospol$}
  \end{prooftree} 
          \end{minipage} 
          \begin{prooftree}
    \AxiomC{$\clause{C} := \clause{C}^\prime \lor [F_{\overline{\tau_j} \to \tau} \; \overline{s^{1,j}}^{1 \leq j \leq n} \simeq t^1_\tau]^\negpol \lor \cdots \lor [F \; \overline{s^{m,j}}^{1 \leq j \leq n} \simeq t^m_\tau]^\negpol$}
    \RightLabel{\scriptsize (FS)}
    \UnaryInfC{$\clause{C}\Big\{F/\lambda \overline{X^j_{\tau_j}}.\, \epsilon Z_\tau.\; \bigwedge_{k=1}^{m} \Big(\big(\bigwedge_{j=1}^{n} X^j = s^{k,j}\big) \longrightarrow Z = t^k\Big)\Big\}$}
  \end{prooftree} 
        \end{minipage}
      }
      \caption{Examples of Leo-III's calculus rules. \textit{Technical
        preliminaries}:  $s \simeq t$ denotes an equation of HOL terms, where $\simeq$ is assumed to be symmetric.
  A literal $\ell$ is a signed equation, written $[s \simeq t]^\alpha$
  where $\alpha \in \{\pospol,\negpol\}$ is the polarity of
  $\ell$. Literals of form $[s_o]^\alpha$ 
  are a shorthand for $[s_o \simeq \syntrue]^\alpha$.
  A clause $\clause{C}$ is a multiset of literals, denoting its disjunction.
  For brevity, if $\clause{C},\clause{D}$ are 
  clauses and $\ell$ is a literal, $\clause{C} \lor \ell$
  and $\clause{C} \lor \clause{D}$ denote the multi-union 
  $\clause{C} \cup \{ \ell \}$ and $\clause{C} \cup \clause{D}$, respectively.
  $s|_\pi$ is the subterm of $s$ at position $\pi$, and $s[r]_\pi$ denotes the term
   that is created by replacing the subterm of $s$ at position $\pi$
   by $r$. \label{fig:calculus}}    
    \end{figure}

  \begin{description}
  \item{\emph{Clause normalization.}}
    Leo-III employs \emph{definitional clausification} to reduce the number of
    clauses. Moreover, \emph{miniscoping} is employed prior to clausification.
    Further normalization rules are straight-forward. 
  \item{\emph{Primary inferences.}}
  The primary inference rules of Leo-III are \emph{paramodulation} (Para),
  \emph{equality factoring} (EqFac) and \emph{primitive substitution} (PS)
  as displayed in Fig.~\ref{fig:calculus}.
  The first two introduce \emph{unification constraints} that are encoded as negative
  literals. Note that these rules are unordered and produce numerous redundant
   clauses. Leo-III uses several heuristics to restrict the number of inferences,
   including a \emph{HO term ordering}. 
   While these restrictions sacrifice completeness in general, recent evaluations
   confirm practicality of this approach (cf. evaluation
   in~\S\ref{sec:eval}); complete search may be retained
   though.
  PS instantiates free variables at top-level
  with approximations of predicate formulas using so-called \emph{general bindings}
  $\mathcal{GB}_\tau^{C}$~\cite[\S2]{J30}.
  \item{\emph{Unification.}}
  Unification in Leo-III uses a variant of \emph{Huet's pre-unification rules}. Negative
  equality literals are interpreted as unification constraints and are attempted to
  be \emph{solved eagerly} by unification. In contrast to LEO-II, Leo-III uses \emph{pattern
  unification} whenever possible. In order to ensure termination, the
  pre-unification \emph{search is limited} to a configurable depth.
  \item{\emph{Extensionality rules.}}
  Dedicated \emph{extensionality rules} are used in order to eliminate the need for
  extensionality axioms in the search space. The rules are similar to those of
  LEO-II~\cite{J30}.
  \item{\emph{Clause contraction.}}
  In addition to standard simplification routines, Leo-III implements 
  are variety of (equational) \emph{simplification procedures}, including \emph{subsumption},
  destructive \emph{equality resolution}, heuristic \emph{rewriting} and contextual \emph{unit cutting}
  (simplify-reflect).
  \item{\emph{Defined Equalities.}}
  Leo-III scans for common definitions of equality predicates and heuristically
  instantiates (or replaces) them with \emph{primitive equality}.
  \item{\emph{Choice.}}
  Leo-III is designed for HOL with choice ($\epsilon^\tau_{(\tau\to o)\to\tau}$
  being a choice operator for type $\tau$).
  Rule Choice \emph{instantiates choice predicates} for
  subterms that represent either concrete choice operator applications (if $E \equiv \epsilon$)
  or potential applications of choice (if $E$ is a free variable of the clause).
  \item{\emph{Function synthesis.}}
  If plain unification fails for a set of unification constraints,
  Leo-III may try to \emph{synthesise function specifications} by rule FS using
  special choice instances that simulate suitable if-then-else terms.
  In general, this rule tremendously increases the search
  space. However, it also enables Leo-III to solve some hard problems
  (with TPTP rating 1.0).
  Also, Leo-III supports improved \emph{reasoning with injective functions} by postulating
  the existence of left-inverses, cf. rule INJ.
  \item{\emph{Heuristic instantiation.}}
  Prior to clause normalization, Leo-III might \emph{instantiate universally quantified}
  variables. This include exhaustive instantiation of finite types as well as
  partial instantiation for otherwise interesting types.
  \end{description}

\section{Modal Logic Reasoning}
\newcommand{\nec}{\Box} \newcommand{\pos}{\Diamond} Modal logics have
many relevant applications in computer science, artificial
intelligence, mathematics and computational linguistics.  They also
play an important role in many areas of philosophy, including
ontology, ethics, philosophy of mind and philosophy of science. Many
challenging applications, as recently explored in metaphysics, require
FO or HO modal
logics (HOMLs). The development of ATPs
for these logics, however, is still in its infancy. 


Leo-III is addressing this gap. In addition to its HOL reasoning capabilities,
it is the first ATP that natively
supports a very wide range of normal HOMLs. To achieve this, Leo-III
internally implements the shallow semantical embeddings approach
\cite{J21,C62}. The key idea in this approach is to provide and exploit 
faithful mappings for HOML input problems to HOL that
encode its Kripke-style semantics.
An example is as follows:

\begin{description}
\item{A} The user inputs a HOML problem in a suitably adapted TPTP syntax, e.g. \\
{\small 
 \verb|thf(1,conjecture,( ! [P:$i>$o,F:$i>$i, X:$i]: (? [G:$i>$i]:| \\
 \verb|      (($dia @ ($box @ (P @ (F @ X)))) => ($box @ (P @ (G @ X))))))).|\\
}
which encodes $\forall P_{\iota\rightarrow o}\forall F_{\iota\to\iota}\forall
X_\iota \exists G_{\iota\rightarrow \iota} (\pos \nec P(F(X)) \Rightarrow \nec
P(G(X)))$, with \verb|$box| \sloppy and 
  \verb|$dia| representing the (mono-)modal operators. This example
  formula (an instance of a corollary of Becker's postulate) is valid 
  in \textbf{S5}.
\item{B} In the header of the input file the user specifies the logic of interest,
  say modal-logic
  \textbf{S5} with rigid constants, constant domain quantifiers and a
  global consequence relation. For this purpose the TPTP language has
  been suitably extended:\footnote{Cf.
    \url{http://www.cs.miami.edu/~tptp/TPTP/Proposals/LogicSpecification.html}
    for the latest TPTP language proposal for modal logics as implemented in Leo-III.} 

{\small
 \verb| thf(simple_s5, logic, ($modal := [|  \\
 \verb|       $constants := $rigid, $quantification := $constant,| \\
 \verb|       $consequence := $global, $modalities := $modal_system_S5 ])).|
}

\item{C} When being called with this input file, Leo-III parses and analyses
  it, automatically selects and unfolds the corresponding definitions
  of the semantical embedding approach, adds
  appropriate axioms and then starts reasoning in (meta-logic)
  HOL. Subsequently, it returns SZS 
  compliant result information and, if successful, also a proof
  object just as for standard HOL problems.
  Leo-III's proof for the embedded problem is verified by GDV~\cite{DBLP:journals/jar/Sutcliffe17}
  in \SI{356}{\second}.
\end{description}
  
As of version 1.2, Leo-III supports (but is not limited to) FO and HO extensions of the well known modal
logic cube. When taking the different parameter combinations into account
(constant/cumulative/varying domain semantics, rigid/non-rigid
constants, local/global consequence relation, and further semantical parameter)  this amounts to
more than 120 supported HOMLs.\footnote{Cf.~\cite[\S 2.2]{C62}; we refer to
  the literature~\cite{blackburn2006handbook} for more details on HOML.} 
The exact number of supported logics is in fact
much higher, since Leo-III also supports multi-modal logics
with independent modal system specification for each modality.
Also, user-defined combinations of rigid and non-rigid constants 
and different quantification semantics per type domain are possible.
\textit{Related provers} are in contrast limited to
propositional logics or support a small range of FO modal
logics only~\cite{MSPASS,METTEL2,C34}.
In the restricted logic settings of the related systems, the
embedding approach
used by Leo-III is still competitive; cf. evaluation below.

\section{Evaluation\label{sec:eval}}
  In order to quantify the performance of Leo-III, an
  evaluation based on various benchmarks was conducted. 
  Three benchmark data sets were used:
  \begin{itemize}
    \item \emph{TPTP TH0} (2463 problems) is the set of all monomorphic HOL (TH0)
          problems from the TPTP library v7.0.0~\cite{DBLP:journals/jar/Sutcliffe17}
          that are annotated as theorems.
          The TPTP problem library is a de-facto standard for the evaluation of ATP systems.
    \item \emph{TPTP TH1} (442 problems) is the subset of all 666 polymorphic
          HOL (TH1) problems from TPTP v7.0.0 that are annotated
          as theorems and do not contain arithmetic.
          The problems mainly consist of HOL Light core exports and
          Sledgehammer translations of various Isabelle theories.
    \item \emph{QMLTP} (580 problems) is the subset of
          all mono-modal benchmarks from the QMLTP library
          1.1~\cite{qmltp}. The QMLTP library only contains
          propositional and FO modal logic problems.
          Since each problem may have a different validity status for each
          semantics of modal logic, all problems (and not only those marked as
          theorem) are selected. The total number of tested problems
          thus is 580 (raw problems) $\times$ 5 (logics) $\times$ 3
          (domain  conditions). QMLTP assumes rigid constant symbols 
          and a local consequence relation.

  \end{itemize}
 
  \noindent The evaluation measurements were taken on the StarExec cluster
  in which each compute node is a \SI{64}{\bit} Red Hat Linux (kernel 3.10.0) machine featuring
  \SI{2.40}{\giga\hertz} Intel Xeon quad-core processors and a main memory of \SI{128}{\giga\byte}.
  For each problem, every prover was given a CPU time limit of \SI{240}{\second}.
  The following theorem provers were employed in one or more of the
  experiments:
  Isabelle/HOL 2016 (TH0/TH1)~\cite{nipkow2002isabelle}, Satallax 3.0 (TH0)~\cite{Satallax},
  Satallax 3.2 (TH0), LEO-II 1.7.0 (TH0),
  Leo-III 1.2 (TH0/TH1/QMLTP), Zipperposition 1.1 (TH0) and
  MleanCoP 1.3~\cite{otten2014mleancop}
  (QMLTP).
  
  The experimental results are discussed next: 
\begin{table}[tb]
  \centering
  \caption{Detailed result of the benchmark measurements\label{table:leo:eval}}
  \subfloat[TPTP TH0 data set (2463 problems)]{
    \resizebox{\linewidth}{!}{
    \begin{tabular}{l|rr|rrrr|rr|rr}
    \textbf{Systems}       & \multicolumn{2}{c|}{\textbf{Solved}}               & \multicolumn{4}{c|}{\textbf{SZS Results}}                                                                                & \multicolumn{2}{c|}{\textbf{Avg. Time} [\si{\second}]}         & \multicolumn{2}{c}{\textbf{$\Sigma$ Time} [\si{\second}]} \\
    \textbf{}              & \multicolumn{1}{l|}{\textbf{Abs.}} & \textbf{Rel.} & \multicolumn{1}{l|}{\textbf{THM}} & \multicolumn{1}{l|}{\textbf{CAX}} & \multicolumn{1}{l|}{\textbf{GUP}} & \textbf{TMO} & \multicolumn{1}{r|}{\;\;\textbf{CPU}} & \textbf{WC} &\multicolumn{1}{l|}{\textbf{CPU}}     & \textbf{WC}     \\ \hline
    \textbf{Satallax 3.2}  & 2140                               & 86.89         & 2140                              & 0                                 & 2                                 & 321          & 12.26                             & 12.31       & 26238            & 26339           \\
    \textbf{Leo-III}              & 2053                               & 83.39         & 2045                              & 8                                 & 15                                & 394          & 15.39                             & 5.61        & 31490            & 11508           \\
    \textbf{Satallax 3.0}  & 1972                               & 80.06         & 2028                              & 0                                 & 2                                 & 433          & 17.83                             & 17.89       & 36149            & 36289           \\
    \textbf{LEO-II}        & 1788                               & 72.63         & 1789                              & 0                                 & 43                                & 603          & 5.84                              & 5.96        & 10452            & 10661           \\
    \textbf{Zipperposition}     & 1318                               & 53.51         & 1318                              & 0                                 & 360                               & 785          & 2.60                               & 2.73        & 3421             & 3592            \\
    \textbf{Isabelle/HOL} & 0                                  & 0.00             & 2022                              & 0                                 & 1                                 & 440          & 46.46                             & 33.44       & 93933            & 67610          
    \end{tabular}
    }
  }
  \\
  \subfloat[TPTP TH1 data set (442 problems)]{
    \resizebox{\linewidth}{!}{
    \begin{tabular}{l|rr|rrrr|rr|rr}
    \textbf{Systems}       & \multicolumn{2}{c|}{\textbf{Solved}}               & \multicolumn{4}{c|}{\textbf{SZS Results}}                                                                                & \multicolumn{2}{c|}{\textbf{Avg. Time} [\si{\second}]}         & \multicolumn{2}{c}{\textbf{$\Sigma$ Time} [\si{\second}]} \\
    \textbf{}              & \multicolumn{1}{l|}{\textbf{Abs.}} & \textbf{Rel.} & \multicolumn{1}{l|}{\textbf{THM}} & \multicolumn{1}{l|}{\textbf{CAX}} & \multicolumn{1}{l|}{\textbf{GUP}} & \textbf{TMO} & \multicolumn{1}{r|}{\;\;\textbf{CPU}} & \textbf{WC} & \multicolumn{1}{l|}{\textbf{CPU}}    & \textbf{WC}     \\ \hline
    \textbf{Leo-III}  & 185 &41.86& 183 & 2 & 8 & 249 & 49.18 & 24.93 & 9099&  4613\\
    \textbf{Isabelle/HOL} & 0 &0.00& 237 & 0 & 23 & 182 & 93.53 &81.44 & 22404& 19300
    \end{tabular}
    }
  }
    
    \end{table}
    
    \begin{figure}[tb] 
	  \centering
	  \caption{Comparison of Leo-III and MleanCoP on the QMLTP data set (580 problems)\label{fig:qmltp}}
	  \begin{tikzpicture} 
		  \begin{axis}[
			  ybar=0pt,
			  xmin = 0.5,
			  xmax = 12.5,
			  ymin = 0,
			  ymax = 470,
			  axis x line* = bottom,
			  axis y line* = left,
			  ylabel= \# solved problems,
			  width= .95\textwidth,
			  height = 0.45\textwidth,
			  ymajorgrids = true,
			  extra y ticks={100,300},
			  bar width = 3mm,
			  xtick = {1,2,3,4,5,6,7,8,9,10,11,12},
			  xticklabels = {D/vary, D/cumul, D/const, T/vary, T/cumul, T/const, S4/vary, S4/cumul, S4/const, S5/vary, S5/cumul, S5/const},
			  x tick label style={rotate=45, anchor=east},
			  legend style={at={(0.03,0.97)},anchor=north west}
			  ]
			  \addplot+[blue_1, thick,draw=none,no markers] coordinates {
			    (1,159)
			    (2,178)
			    (3,203)
			    (4,209)
			    (5,233)
			    (6,263)
			    (7,246)
			    (8,280)
			    (9,316)
			    (10,340)
			    (11,457)
			    (12,457)
			  };
			  
			  \addplot+[blue_2b, thick,draw=none,no markers] coordinates { 
			    (1,185)
			    (2,206)
			    (3,223)
			    (4,224)
			    (5,251)
			    (6,271)
			    (7,289)
			    (8,350)
			    (9,366)
			    (10,360)
			    (11,436)
			    (12,436)
			  };
			  \legend{Leo-III, MleanCoP}
		  \end{axis} 
	  \end{tikzpicture}
  \end{figure}
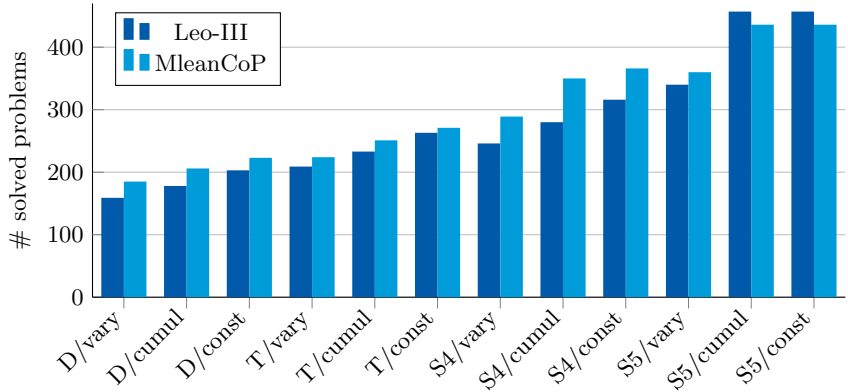
    
  \textit{TPTP TH0.} Table~\ref{table:leo:eval} (a) displays each system's performance on the TPTP TH0 data set.
  For each system the absolute number (Abs.) and relative share (Rel.) of solved problems is
  displayed. Solved here means that a system is able to establish the SZS status \texttt{Theorem}
  and also emits a proof certificate that substantiates this claim.
  All results of the system, whether successful or not, are counted and categorized
  as THM (\texttt{Theorem}), CAX (\texttt{ContradictoryAxioms}), GUP (\texttt{GaveUp}) and TMO (\texttt{TimeOut})
  for the respective SZS status of the returned result.
  Additionally, the average and sum of all CPU times
  and wall clock (WC) times over all solved problems is presented.
  
  Leo-III successfully solves 2053 of 2463 problems
  (roughly \SI{83.39}{\percent})
  from the TPTP TH0 data set. This is 735 (\SI{35.8}{\percent}) more than Zipperposition, 
  264 (\SI{12.86}{\percent}) more than LEO-II and 81 (\SI{3.95}{\percent}) more than Satallax 3.0.
  The only ATP system that solves more problems is the most recent version of Satallax
  (3.2) that successfully solves 2140 problems, which is approximately
  \SI{4.24}{\percent} more than Leo-III. 
  Isabelle currently does not emit proof certificates (hence zero solutions).
  Even if results without explicit proofs are counted,
  Leo-III would still have a slightly higher number of problems solved than Satallax 3.0 and
  Isabelle/HOL with 25 (\SI{1.22}{\percent}) and 31 (\SI{1.51}{\percent}) additional solutions, respectively.
  Leo-III, Satallax (3.2), Zipperposition and LEO-II produce 18,
  17, 15 and 3 unique solutions, respectively. Evidently, Leo-III currently
  produces more unique solutions than any other ATP system in this setting.
  %
  Leo-III
  solves nine previously unsolved problems and three problems that are currently
  not solved by any other system.\footnote{
  This information is extracted from the TPTP
  problem rating information that is attached to each problem.
  The exact, previously unsolved, problems are \texttt{NLP004\textasciicircum7},
    \texttt{SET013\textasciicircum7}, \texttt{SEU558\textasciicircum1}, \texttt{SEU683\textasciicircum1},
    \texttt{SEV143\textasciicircum5},
    \texttt{SYO037\textasciicircum1}, \texttt{SYO062\textasciicircum4.004},
    \texttt{SYO065\textasciicircum4.001}
    and \texttt{SYO066\textasciicircum4.004}. 
    The three problems that are currently not solved by any other ATP system are
    \texttt{MSC007\textasciicircum1.003.004}, \texttt{SEU938\textasciicircum5} and 
    \texttt{SEV106\textasciicircum5}.
  }
  
  Satallax, LEO-II and Zipperposition show only small differences between
  their individual CPU and WC time on average and sum. A more precise measure for a system's utilization of 
  multiple cores is the so-called \emph{core usage}. It is given by the
  average of the ratios of used CPU time to used wall clock time over all solved problems.
  The core usage of Leo-III for the TPTP TH0 data set is roughly $2.52$. This means that,
  on average, two to three CPU cores are used during proof search by Leo-III.
  Satallax (3.2), LEO-II and Zipperposition show a quite opposite behavior with
  core usages of $0.64$, $0.56$ and $0.47$, respectively.
  
  \textit{TPTP TH1.}
  Currently, there exist only few ATP systems that are capable of reasoning
  within polymorphic HOL as specified by TPTP TH1.
  The only exceptions are
  HOL(y)Hammer
  and Isabelle/HOL
  that schedule proof tactics within HOL Light and Isabelle/HOL,
  respectively.
  Unfortunately, only Isabelle/HOL was available for instrumentation
  in a reasonably recent and stable
  version.
  Table~\ref{table:leo:eval} (b) displays the measurement results for the TPTP TH1 data
  set. When disregarding proof certificates, Isabelle/HOL finds 237 theorems (\SI{53.62}{\percent})
  which is roughly \SI{28.1}{\percent} more than the number of solutions founds by Leo-III.
  Leo-III and Isabelle/HOL produce 35 and 69 unique solutions, respectively.
  
  \textit{QMLTP.} 
  For each semantical setting supported by MleanCoP, which is the
  strongest FO modal logic prover available to date~\cite{C34}, the number of theorems found 
  by both Leo-III and MleanCoP in the QMLTP data set is presented in Fig.~\ref{fig:qmltp}.
  Leo-III is fairly competitive with MleanCoP (weaker by maximal 
  \SI{14.05}{\percent}, minimal \SI{2.95}{\percent} and
  \SI{8.90}{\percent} on average) for all \textbf{D} and \textbf{T}
  variants.  For all \textbf{S4} variants, the gap between both
  systems increases (weaker by maximal \SI{20.00}{\percent}, minimal
  \SI{13.66}{\percent} and \SI{16.18}{\percent} on average). For
  \textbf{S5} variants, Leo-III is very effective (stronger by
  \SI{1.36}{\percent} on average) and it is ahead of MleanCoP for
  \textbf{S5}/{const} and \textbf{S5}/{cumul} (which coincide).  This is due to
  the encoding of the \textbf{S5} accessibility relation in Leo-III 1.2 as
  the universal relation between possible worlds as opposed to its prior 
  encoding (cf.~\cite{C34,C62}) as an equivalence relation.
  Leo-III contributes 199 solutions to previously unsolved problems.

  For HOML there exist no competitor systems
  with which Leo-III could be compared. 
  
\section{Summary}
Leo-III is a state-of-the-art higher-order reasoning system offering
many relevant features and capabilities. Due to its wide range of
natively supported classical and non-classical logics, which include
polymorphic HO logic and numerous FO and HO modal logics, the system
has many topical applications in computer science, AI, maths and
philosophy.
Additionally, an evaluation on heterogeneous benchmark sets shows that
Leo-III is also one of the most effective HO ATP systems to date.
Leo-III complies with existing {TPTP/TSTP} standards, gives
detailed proof certificates and plays a pivotal role in the ongoing extension of the TPTP
library and infrastructure to support modal logic reasoning.


\bibliographystyle{splncs}
\bibliography{main,chris}

\begin{thebibliography}{10}

\bibitem{J30}
Benzm{\"u}ller, C., Sultana, N., Paulson, L.C., Thei{\ss}, F.:
\newblock The higher-order prover {LEO-II}.
\newblock Journal of Automated Reasoning \textbf{55}(4) (2015)  389--404

\bibitem{J22}
Sutcliffe, G., Benzm{\"u}ller, C.:
\newblock Automated reasoning in higher-order logic using the {TPTP THF}
  infrastructure.
\newblock Journal of Formalized Reasoning \textbf{3}(1) (2010)  1--27

\bibitem{C45short}
Wisniewski, M., Steen, A., Benzm{\"{u}}ller, C.:
\newblock {LeoPARD} - {A} generic platform for the implementation of
  higher-order reasoners.
\newblock In Kerber, M.,  et~al., eds.: Intelligent Computer Mathematics.
  Volume 9150 of LNCS., Springer (2015)  325--330

\bibitem{Schulz:2002:EBT:1218615.1218621}
Schulz, S.:
\newblock E - a brainiac theorem prover.
\newblock AI Commun. \textbf{15}(2,3) (August 2002)  111--126

\bibitem{DBLP:journals/jar/Sutcliffe17}
Sutcliffe, G.:
\newblock The {TPTP} problem library and associated infrastructure - from {CNF}
  to {TH0}, {TPTP} v6.4.0.
\newblock J. Autom. Reasoning \textbf{59}(4) (2017)  483--502

\bibitem{DBLP:conf/cade/BlanchetteP13}
Blanchette, J.C., Paskevich, A.:
\newblock {TFF1:} the {TPTP} typed first-order form with rank-1 polymorphism.
\newblock In Bonacina, M.P., ed.: Automated Deduction -- {CADE-24}. Volume 7898
  of LNCS., Springer (2013)  414--420

\bibitem{DBLP:conf/cade/KaliszykSR16}
Kaliszyk, C., Sutcliffe, G., Rabe, F.:
\newblock {TH1:} the {TPTP} typed higher-order form with rank-1 polymorphism.
\newblock In Fontaine, P.,  et~al., eds.: 5th PAAR Workshop. Volume 1635 of
  {CEUR} Workshop Proceedings., CEUR-WS.org (2016)  41--55

\bibitem{blackburn2006handbook}
Blackburn, P., van Benthem, J.F., Wolter, F.:
\newblock Handbook of modal logic. Volume~3.
\newblock Elsevier (2006)

\bibitem{qmltp}
Raths, T., Otten, J.:
\newblock {The {QMLTP} Problem Library for First-Order Modal Logics}.
\newblock In Gramlich, B.,  et~al., eds.: {IJCAR} 2012. Volume 7364 of LNCS.,
  Springer (2012)  454--461

\bibitem{sep-type-theory-church}
Andrews, P.:
\newblock Church's type theory.
\newblock In Zalta, E.N., ed.: The Stanford Encyclopedia of Philosophy.
\newblock Metaphysics Research Lab, Stanford University (2014)

\bibitem{B5}
Benzm{\"u}ller, C., Miller, D.:
\newblock Automation of higher-order logic.
\newblock In Gabbay, D.M., Siekmann, J.H., Woods, J., eds.: Handbook of the
  History of Logic, Volume 9 --- Computational Logic.
\newblock North Holland, Elsevier (2014)  215--254

\bibitem{J21}
Benzm{\"u}ller, C., Paulson, L.:
\newblock Multimodal and intuitionistic logics in simple type theory.
\newblock The Logic Journal of the IGPL \textbf{18}(6) (2010)  881--892

\bibitem{C62}
Glei{\ss}ner, T., Steen, A., Benzm{\"u}ller, C.:
\newblock Theorem provers for every normal modal logic.
\newblock In Eiter, T., Sands, D., eds.: LPAR-21. Volume~46 of EPiC Series in
  Computing., Maun, Botswana, EasyChair (2017)  14--30

\bibitem{MSPASS}
Hustadt, U., Schmidt, R.A.:
\newblock Mspass: Modal reasoning by translation and first-order resolution.
\newblock In: TABLEAUX. Volume 1847., Springer (2000)  67--71

\bibitem{METTEL2}
Tishkovsky, D., Schmidt, R.A., Khodadadi, M.:
\newblock {The tableau prover generator MetTeL2}.
\newblock In: European Workshop on Logics in AI, Springer (2012)  492--495

\bibitem{C34}
Benzm{\"u}ller, C., Otten, J., Raths, T.:
\newblock Implementing and evaluating provers for first-order modal logics.
\newblock In Raedt, L.D.,  et~al., eds.: ECAI 2012. Volume 242 of Frontiers in
  AI and Applications., Montpellier, France, IOS Press (2012)  163--168

\bibitem{nipkow2002isabelle}
Nipkow, T., Paulson, L.C., Wenzel, M.:
\newblock Isabelle/HOL: A Proof Assistant for Higher-Order Logic.
\newblock Lecture Notes in Computer Science. Springer (2002)

\bibitem{Satallax}
Brown, C.E.:
\newblock Satallax: An automatic higher-order prover.
\newblock In: Automated Reasoning. Volume 7364 of LNCS.
\newblock Springer Berlin Heidelberg (2012)  111--117

\bibitem{otten2014mleancop}
Otten, J.:
\newblock Mleancop: A connection prover for first-order modal logic.
\newblock In: International Joint Conference on Automated Reasoning, Springer
  (2014)  269--276

\end{thebibliography}

\appendix
\renewcommand{\ttdefault}{pcr}
\lstset{basicstyle=\ttfamily\scriptsize,frame=single,breaklines=true,numbers=none,morekeywords={thf,., definition, logic, axiom, conjecture, type, plain, negated_conjecture},morecomment=[l]\%}
\section{Reasoning in HOML with Leo-III: Corollary of Becker's Postulate}
This appendix will be removed from the final version of the paper. We included it here as potentially useful additional information for interested reviewers.
\subsection*{Problem statement}
The original problem statement formulated in modal THF syntax is given by:
\begin{lstlisting}
thf(simple_s5,logic,(
    $modal := 
      [ $constants := $rigid, 
        $quantification := $constant, 
        $consequence := $global, 
        $modalities := $modal_system_S5 ] )).

thf(1,conjecture,(
    ! [P: ( $i > $o ),F: ( $i > $i ),X: $i] :
    ? [Q: ( $i > $i )] :
      ( ( $dia @ ( $box @ ( P @ ( F @ X ) ) ) )
     => ( $box @ ( P @ ( Q @ X ) ) ) ) )).
\end{lstlisting}

\subsection*{Internal problem representation}
Internally, Leo-III faithfully maps the problem into
in classical HOL (THF). The TPTP representation of this internally
generated problem is as follows:
\begin{lstlisting}
% -------------------------------------------------------------------------
% modal definitions 
% -------------------------------------------------------------------------
% declare type for possible worlds
thf(mworld_type,type,(
    mworld: $tType )).

% define valid operator
thf(mvalid_type,type,(
    mvalid: ( mworld > $o ) > $o )).

thf(mvalid_def,definition,
    ( mvalid
    = ( ^ [S: ( mworld > $o )] :
        ! [W: mworld] :
          ( S @ W ) ) )).

% define nullary, unary and binary connectives which are no quantifiers
thf(mimplies_type,type,(
    mimplies: ( mworld > $o ) > ( mworld > $o ) > mworld > $o )).

thf(mimplies,definition,
    ( mimplies
    = ( ^ [A: ( mworld > $o ),B: ( mworld > $o ),W: mworld] :
          ( ( A @ W )
         => ( B @ W ) ) ) )).

thf(mdia_type,type,(
    mdia: ( mworld > $o ) > mworld > $o )).

thf(mdia_def,definition,
    ( mdia
    = ( ^ [A: ( mworld > $o ),W: mworld] :
        ? [V: mworld] :
          ( A @ V ) ) )).

thf(mbox_type,type,(
    mbox: ( mworld > $o ) > mworld > $o )).

thf(mbox_def,definition,
    ( mbox
    = ( ^ [A: ( mworld > $o ),W: mworld] :
        ! [V: mworld] :
          ( A @ V ) ) )).

% define exists quantifiers
thf(mexists_const_type__o__d_i_t__d_i_c_,type,(
    mexists_const__o__d_i_t__d_i_c_: ( ( $i > $i ) > mworld > $o ) > mworld > $o )).

thf(mexists_const__o__d_i_t__d_i_c_,definition,
    ( mexists_const__o__d_i_t__d_i_c_
    = ( ^ [A: ( ( $i > $i ) > mworld > $o ),W: mworld] :
        ? [X: ( $i > $i )] :
          ( A @ X @ W ) ) )).

% define for all quantifiers
thf(mforall_const_type__d_i,type,(
    mforall_const__d_i: ( $i > mworld > $o ) > mworld > $o )).

thf(mforall_const__d_i,definition,
    ( mforall_const__d_i
    = ( ^ [A: ( $i > mworld > $o ),W: mworld] :
        ! [X: $i] :
          ( A @ X @ W ) ) )).

thf(mforall_const_type__o__d_i_t__o_mworld_t__d_o_c__c_,type,(
    mforall_const__o__d_i_t__o_mworld_t__d_o_c__c_: ( ( $i > mworld > $o ) > mworld > $o ) > mworld > $o )).

thf(mforall_const__o__d_i_t__o_mworld_t__d_o_c__c_,definition,
    ( mforall_const__o__d_i_t__o_mworld_t__d_o_c__c_
    = ( ^ [A: ( ( $i > mworld > $o ) > mworld > $o ),W: mworld] :
        ! [X: ( $i > mworld > $o )] :
          ( A @ X @ W ) ) )).

thf(mforall_const_type__o__d_i_t__d_i_c_,type,(
    mforall_const__o__d_i_t__d_i_c_: ( ( $i > $i ) > mworld > $o ) > mworld > $o )).

thf(mforall_const__o__d_i_t__d_i_c_,definition,
    ( mforall_const__o__d_i_t__d_i_c_
    = ( ^ [A: ( ( $i > $i ) > mworld > $o ),W: mworld] :
        ! [X: ( $i > $i )] :
          ( A @ X @ W ) ) )).
% -------------------------------------------------------------------------
% transformed problem
% -------------------------------------------------------------------------
thf(1,conjecture,
    ( mvalid
    @ ( mforall_const__o__d_i_t__o_mworld_t__d_o_c__c_
      @ ^ [P: ( $i > mworld > $o )] :
          ( mforall_const__o__d_i_t__d_i_c_
          @ ^ [F: ( $i > $i )] :
              ( mforall_const__d_i
              @ ^ [X: $i] :
                  ( mexists_const__o__d_i_t__d_i_c_
                  @ ^ [Q: ( $i > $i )] :
                      ( mimplies @ ( mdia @ ( mbox @ ( P @ ( F @ X ) ) ) ) @ ( mbox @ ( P @ ( Q @ X ) ) ) ) ) ) ) ) )).
\end{lstlisting}

\subsection*{Leo-III's proof output}
Leo-III proves the problem in approx. \SI{3}{\second}.
The exact output of Leo-III, including the generated proof certificate, is
given in the following. The proof can be verified by GDV with Isabelle/HOL
as HO ATP back end in \SI{356}{\second}.
\begin{lstlisting}
% [INFO]   Input problem is modal. Running modal-to-HOL transformation from semantics specification contained in the problem file ... 
% [INFO]   Running in sequential loop mode. 
% [INFO]   Parsing finished. Scanning for conjecture ... 
% [INFO]   Found a conjecture and 0 axioms. Running axiom selection ... 
% [INFO]   Axiom selection finished. Selected 0 axioms (removed 0 axioms). 
% [INFO]   Type checking passed. Searching for refutation ... 
% Time passed: 3010ms
% Effective reasoning time: 688ms
% Solved by strategy<name(default),share(1.0),primSubst(1),sos(false),unifierCount(1),uniDepth(8),boolExt(true),choice(true),renaming(true),funcspec(false), domConstr(0),specialInstances(-1),restrictUniAttempts(true)>
% Axioms used in derivation (0): 
% No. of inferences in proof: 9
% No. of processed clauses: 2
% No. of generated clauses: 1
% No. of forward subsumed clauses: 0
% No. of backward subsumed clauses: 0
% No. of ground rewrite rules in store: 0
% No. of non-ground rewrite rules in store: 2
% No. of positive (non-rewrite) units in store: 0
% No. of negative (non-rewrite) units in store: 0
% No. of choice functions detected: 0
% No. of choice instantiations: 0

% SZS status Theorem for becker.p : 3010 ms resp. 688 ms w/o parsing
% SZS output start CNFRefutation for becker.p
thf(mworld_type,type,( mworld: $tType )).
thf(mvalid_type,type,( mvalid: ( mworld > $o ) > $o )).
thf(mvalid_def,definition,
    ( mvalid
    = ( ^ [A: ( mworld > $o )] :
        ! [B: mworld] :
          ( A @ B ) ) )).
thf(mimplies_type,type,(
    mimplies: ( mworld > $o ) > ( mworld > $o ) > mworld > $o )).
thf(mimplies_def,definition,
    ( mimplies
    = ( ^ [A: ( mworld > $o ),B: ( mworld > $o ),C: mworld] :
          ( ( A @ C )
         => ( B @ C ) ) ) )).
thf(mdia_type,type,( mdia: ( mworld > $o ) > mworld > $o )).
thf(mdia_def,definition,
    ( mdia
    = ( ^ [A: ( mworld > $o ),B: mworld] :
        ? [C: mworld] :
          ( A @ C ) ) )).
thf(mbox_type,type,( mbox: ( mworld > $o ) > mworld > $o )).
thf(mbox_def,definition,
    ( mbox
    = ( ^ [A: ( mworld > $o ),B: mworld] :
        ! [C: mworld] :
          ( A @ C ) ) )).
thf(mexists_const__o__d_i_t__d_i_c__type,type,(
    mexists_const__o__d_i_t__d_i_c_: ( ( $i>$i ) > mworld > $o ) > mworld > $o )).
thf(mexists_const__o__d_i_t__d_i_c__def,definition,
    ( mexists_const__o__d_i_t__d_i_c_
    = ( ^ [A: ( ( $i > $i ) > mworld > $o ),B: mworld] :
        ? [C: ( $i > $i )] :
          ( A @ C @ B ) ) )).
thf(mforall_const__d_i_type,type,(
    mforall_const__d_i: ( $i > mworld > $o ) > mworld > $o )).
thf(mforall_const__d_i_def,definition,
    ( mforall_const__d_i
    = ( ^ [A: ( $i > mworld > $o ),B: mworld] :
        ! [C: $i] :
          ( A @ C @ B ) ) )).
thf(mforall_const__o__d_i_t__o_mworld_t__d_o_c__c__type,type,(
    mforall_const__o__d_i_t__o_mworld_t__d_o_c__c_: ( ( $i > mworld > $o ) > mworld > $o ) > mworld > $o )).
thf(mforall_const__o__d_i_t__o_mworld_t__d_o_c__c__def,definition,
    ( mforall_const__o__d_i_t__o_mworld_t__d_o_c__c_
    = ( ^ [A: ( ( $i > mworld > $o ) > mworld > $o ),B: mworld] :
        ! [C: ( $i > mworld > $o )] :
          ( A @ C @ B ) ) )).
thf(mforall_const__o__d_i_t__d_i_c__type,type,(
    mforall_const__o__d_i_t__d_i_c_: ( ( $i>$i ) > mworld > $o ) > mworld > $o )).
thf(mforall_const__o__d_i_t__d_i_c__def,definition,
    ( mforall_const__o__d_i_t__d_i_c_
    = ( ^ [A: ( ( $i > $i ) > mworld > $o ),B: mworld] :
        ! [C: ( $i > $i )] :
          ( A @ C @ B ) ) )).
thf(sk1_type,type,( sk1: $i > mworld > $o )).
thf(sk2_type,type,( sk2: $i > $i )).
thf(sk3_type,type,( sk3: $i )).
thf(sk4_type,type,( sk4: ( $i > $i ) > mworld )).

thf(1,conjecture,
    ( mvalid
    @ ( mforall_const__o__d_i_t__o_mworld_t__d_o_c__c_
      @ ^ [A: ( $i > mworld > $o )] :
          ( mforall_const__o__d_i_t__d_i_c_
          @ ^ [B: ( $i > $i )] :
              ( mforall_const__d_i
              @ ^ [C: $i] :
                  ( mexists_const__o__d_i_t__d_i_c_
                  @ ^ [D: ( $i > $i )] :
                      ( mimplies @ ( mdia @ ( mbox @ ( A @ ( B @ C ) ) ) ) @ ( mbox @ ( A @ ( D @ C ) ) ) ) ) ) ) ) ),
    file('becker.p',1)).
thf(2,negated_conjecture,(
    ~ ( mvalid
      @ ( mforall_const__o__d_i_t__o_mworld_t__d_o_c__c_
        @ ^ [A: ( $i > mworld > $o )] :
            ( mforall_const__o__d_i_t__d_i_c_
            @ ^ [B: ( $i > $i )] :
                ( mforall_const__d_i
                @ ^ [C: $i] :
                    ( mexists_const__o__d_i_t__d_i_c_
                    @ ^ [D: ( $i > $i )] :
                        ( mimplies @ ( mdia @ ( mbox @ ( A @ ( B @ C ) ) ) ) @ ( mbox @ ( A @ ( D @ C ) ) ) ) ) ) ) ) ) ),
    inference(neg_conjecture,[status(cth)],[1])).
thf(3,plain,(
    ~ ( ! [A: ( $i > mworld > $o ),B: ( $i > $i ),C: $i] :
        ? [D: ( $i > $i )] :
          ( ! [E: mworld] :
              ( A @ ( B @ C ) @ E )
         => ! [E: mworld] :
              ( A @ ( D @ C ) @ E ) ) ) ),
    inference(defexp_and_simp_and_etaexpand,[status(thm)],[2,mvalid_def,mforall_const__o__d_i_t__o_mworld_t__d_o_c__c__def,mforall_const__o__d_i_t__d_i_c__def,mforall_const__d_i_def,mexists_const__o__d_i_t__d_i_c__def,mimplies_def,mdia_def,mbox_def])).
thf(4,plain,(
    ~ ( ! [A: ( $i > mworld > $o ),B: ( $i > $i ),C: $i] :
          ( ! [D: mworld] :
              ( A @ ( B @ C ) @ D )
         => ? [D: ( $i > $i )] :
            ! [E: mworld] :
              ( A @ ( D @ C ) @ E ) ) ) ),
    inference(miniscope,[status(thm)],[3])).
thf(6,plain,(
    ! [A: mworld] :
      ( sk1 @ ( sk2 @ sk3 ) @ A ) ),
    inference(cnf,[status(esa)],[4])).
thf(5,plain,(
    ! [A: ( $i > $i )] :
      ~ ( sk1 @ ( A @ sk3 ) @ ( sk4 @ A ) ) ),
    inference(cnf,[status(esa)],[4])).
thf(7,plain,(
    ! [A: ( $i > $i )] :
      ~ ( sk1 @ ( A @ sk3 ) @ ( sk4 @ A ) ) ),
    inference(simp,[status(thm)],[5])).
thf(8,plain,(
    ! [B: ( $i > $i ),A: mworld] :
      ( ( sk1 @ ( sk2 @ sk3 ) @ A )
     != ( sk1 @ ( B @ sk3 ) @ ( sk4 @ B ) ) ) ),
    inference(paramod_ordered,[status(thm)],[6,7])).
thf(9,plain,( $false ),
    inference(pre_uni,[status(thm)],[8:[bind(A,$thf(sk4 @ ^ [C: $i] : ( sk2 @ sk3 ))),bind(B,$thf(^ [C: $i] : ( sk2 @ sk3 )))]])).
% SZS output end CNFRefutation for becker.p
\end{lstlisting}

\end{document}